\documentclass{article}
\usepackage{ijcai23}

\usepackage{times}
\usepackage{soul}
\usepackage{url}
\usepackage[hidelinks]{hyperref}
\usepackage[utf8]{inputenc}
\usepackage[small]{caption}
\usepackage{graphicx}
\usepackage{amsmath}
\usepackage{amsthm}
\usepackage{booktabs}
\usepackage{algorithm}
\usepackage{algorithmic}
\usepackage{subfigure}
\usepackage{multicol}
\usepackage{multirow}
\usepackage[switch]{lineno}
\usepackage{float}

\title{SAD: Semi-Supervised Anomaly Detection on Dynamic Graphs}

\author{
Sheng Tian$^1$\footnote{Both authors contribute equally.}
\and
Jihai Dong$^{1*}$\and
Jintang Li$^2$\and
Wenlong Zhao$^1$\and
Xiaolong Xu$^1$\and\\
Baokun Wang$^1$\and
Bowen Song$^1$\and
Changhua Meng$^1$\and
Tianyi Zhang$^1$\And
Liang Chen$^2$
\affiliations
$^1$Ant Group\\
$^2$Sun Yat-sen University
\emails
\{tiansheng.ts, dongjihai.djh\}@antgroup.com,
lijt55@mail2.sysu.edu.cn,
\{chicheng.zwl, yiyin.xxl, yike.wbk, bowen.sbw, changhua.mch, zty113091\}@antgroup.com,
chenliang6@mail.sysu.edu.cn
}
\begin{document}

\maketitle

\begin{abstract}

Anomaly detection aims to distinguish abnormal instances that deviate significantly from the majority of benign ones. As instances that appear in the real world are naturally connected and can be represented with graphs, graph neural networks become increasingly popular in tackling the anomaly detection problem. Despite the promising results, research on anomaly detection has almost exclusively focused on static graphs while the mining of anomalous patterns from dynamic graphs is rarely studied but has significant application value. In addition, anomaly detection is typically tackled from semi-supervised perspectives due to the lack of sufficient labeled data. However, most proposed methods are limited to merely exploiting labeled data, leaving a large number of unlabeled samples unexplored. In this work, we present \textbf{semi-supervised anomaly detection (SAD)}, an end-to-end framework for anomaly detection on dynamic graphs. By a combination of a time-equipped \textit{memory bank} and a \textit{pseudo-label contrastive learning module}, SAD is able to fully exploit the potential of large unlabeled samples and uncover underlying anomalies on evolving graph streams. Extensive experiments on four real-world datasets demonstrate that SAD efficiently discovers anomalies from dynamic graphs and outperforms existing advanced methods even when provided with only little labeled data.

\end{abstract}
%\vspace{-0.2cm}

\section{Introduction}

Anomaly detection, which identifies outliers (called \textit{anomalies}) that deviate significantly from normal instances, has been a lasting yet active research area in various research contexts~\cite{ma2021comprehensive}.
In many real-world scenarios, instances are often explicitly connected with each other and can be naturally represented with graphs. Over the past few years, anomaly detection based on graph representation learning has emerged as a critical direction and demonstrated its power in detecting anomalies from instances with abundant relational information. For example, in a financial transaction network where fraudsters would try to cheat normal users and make illegal money transfers, the graph patterns captured from the local and global perspectives can help detect these fraudsters and prevent financial fraud.

In real-world scenarios, it is often challenging to tackle the anomaly detection problem with supervised learning methods since annotated labels in most situations are rare and difficult to acquire. Therefore, current works mainly focus on unsupervised schemes to perform anomaly detection, with prominent examples including autoencoders~\cite{zhou2017anomaly}, adversarial networks~\cite{chen2020generative}, or matrix factorization~\cite{Bandyopadhyay2019outlier}. However, they tend to produce noisy results or uninterested noise instances due to insufficient supervision. Empirically, it is usually difficult to get desired outputs from unsupervised methods without the guidance of precious labels or correct assumptions.

\begin{figure}[t]
    \centering
\includegraphics[width=1\hsize, height=0.3\hsize]{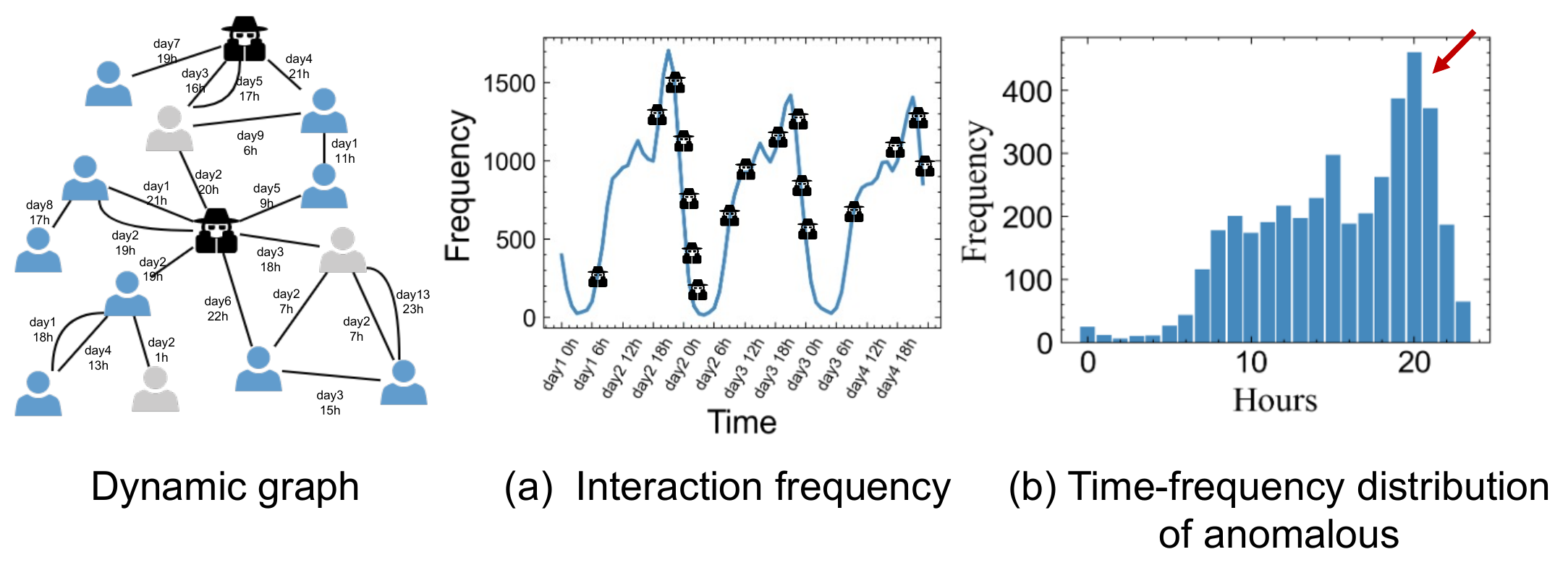}
 \vspace{-0.5cm}
    \caption{Statistical distribution of evolving graph streams on the MOOC dataset. \textbf{Observations}: (a) The frequency of interactions shows a clear cyclical pattern on the time axis. (b) Anomalous samples typically occurred between 8 am and 10 pm, with peaks at around 8 pm.}
	\label{fig:nums_time}
 \vspace{-0.4cm}
\end{figure}

Semi-supervised learning offers a principled way to resolve the issues mentioned above by utilizing few-shot labeled data combined with abundant unlabeled data to detect underlying anomalies~\cite{ma2021comprehensive}. Such methods are practical in real-world applications and can definitely obtain much better performance than fully unsupervised methods when the labeled data is leveraged properly. In recent years, a vast majority of work adopts graph-based semi-supervised learning models for solving anomaly detection problems, resulting in beneficial advances in anomaly analytics techniques~\cite{pang2019deep,ding2021few,qian2021distilling}.

Yet, there are still two fundamental limitations to these existing semi-supervised anomaly detection methods: (1) \textit{Underutilization of unlabeled data}. Although existing semi-supervised approaches have utilized labeled data to perform anomaly detection, few methods take advantage of large-scale unlabeled samples that contain a lot of useful information. As a result, they suffer poor performance particularly when limited labels are provided.
(2) \textit{Lack of dynamic information mining}. In practice, graphs are dynamic in nature, with nodes, edges, and attributes evolving constantly over time. Such characteristics may be beneficial in solving anomaly detection problems. As shown in Figure~\ref{fig:nums_time}, the behavioral statistics of users show certain cyclical patterns in the time domain, which can be directly captured by dynamic graph neural networks. To our best knowledge, most previous attempts mainly focus on static graphs which neglect temporal information, such design has proven to be sub-optimal in dynamic graph scenario~\cite{liu2021taddy}. Although there are some recent efforts~\cite{meng2021semi,liu2021taddy}have explored ways to combine dynamic information in their work, they simply incorporate time as a feature in the model, leading to insufficient temporal information modeling and degraded performance on anomaly detection.

In this work, we propose a novel \textit{semi-supervised anomaly detection (SAD)} framework that tackles the aforementioned limitations. The proposed framework first utilizes a temporal graph network to encode graph data for both labeled and unlabeled samples, followed by an anomaly detector network to predict node anomaly scores. Then, a memory bank is adopted to record each predicted anomaly score together with node time information, to produce a normal sample statistical distribution as prior knowledge and guide the learning of the network subsequently by using a deviation loss. In order to further exploit the potential of large unlabeled data, we introduce a novel pseudo-label contrastive learning module, which leverages the predicted anomaly scores to form pseudo-groups by calculating score distance between nodes - nodes with closer distance will be classified into the same pseudo-group, nodes within the same pseudo-group will form positive pairs for contrastive learning. 

Our main contributions are summarized as follows:
\begin{itemize}
    %\vspace{-0.1cm}
    \item We propose SAD, an end-to-end semi-supervised anomaly detection framework, which is tailored with a time-equipped memory bank and a pseudo-label contrastive learning module to effectively solve the anomaly detection problem on dynamic graphs.
    \item Our proposed SAD framework uses the statistical distribution of unlabeled samples as the reference distribution for loss calculation and generates pseudo-labels correspondingly to participate in supervised learning, which in turn fully exploits the potential of unlabeled samples.
    \item Extensive experiments demonstrate the superiority of the proposed framework compared to strong baselines. More importantly, our proposed approach can also significantly alleviate the label scarcity issue in practical applications.

\end{itemize}

\section{Related Work}
\subsection{Dynamic Graph Neural Networks} 

%2>  附录中增加 不使用节点分类损失的效果。

Graph neural networks (GNNs) have been prevalently used to represent structural information from graph data. However, naive applications of static graph learning paradigms~\cite{william2017inductive,petar2018graph} in dynamic graphs fail to capture temporal information and show up as suboptimal in dynamic graph scenarios~\cite{xu2020tgat}. Thus, some methods~\cite{Singer2019node,Pareja2020evolvegcn} model the evolutionary process of a dynamic graph by discretizing time and reproducing multiple static graph snapshots. Such a paradigm is often infeasible for real-world systems that need to handle each interaction event instantaneously, in which graphs are represented by a continuous sequence of events, i.e., nodes and edges can appear and disappear at any time.

Some recent work has begun to consider the direct modeling of temporal information and proposed a continuous time dynamic graph modeling scheme. For example, DyREP~\cite{Trivedi2019DyRep} uses a temporal point process model, which is parameterized by a recurrent architecture to learn evolving entity representations on temporal knowledge graphs. TGAT~\cite{xu2020tgat} proposes a novel functional time encoding technique with a self-attention layer to aggregate temporal-topological neighborhood features. However, the above methods rely heavily on annotated labels, yet obtaining sufficient annotated labels is usually very expensive, which greatly limits their application in real-world scenarios such as anomaly detection.

\subsection{Anomaly Detection with GNNs} 

Anomalies are rare observations (e.g., data records or events) that deviate significantly from the others in the sample, and graph anomaly detection (GAD)  aims to identify anomalous graph objects (i.e., nodes, edges, or sub-graphs) in a single graph as well as anomalous graphs among a set/database of graphs~\cite{ma2021comprehensive}. The anomaly detection with GNNs is mainly divided into unsupervised and semi-supervised methods. Unsupervised anomaly detection methods are usually constructed with a two-stage task, where a graph representation learning task is first constructed, and then abnormality measurements are applied based on the latent representations~\cite{Akoglu2015graph}. Rader~\cite{li2017radar} detects anomalies by characterizing the residuals of the attribute information and their correlation with the graph structure. Autoencoder~\cite{zhou2017anomaly} learns graph representations through an unsupervised feature-based deep autoencoder model. DOMINANT~\cite{ding2019deep} uses a GCN-based autoencoder framework that uses reconstruction errors from the network structure and node representation to distinguish anomalies. CoLA~\cite{Cola} proposes a self-supervised contrastive learning method based on capturing the relationship between each node and its neighboring substructure to measure the agreement of each instance pair.

A more recent line of work proposes employing semi-supervised learning by using small amounts of labeled samples from the relevant downstream tasks to address the problem that unsupervised learning tends to produce noisy instances. SemiGNN~\cite{wang2019semi} learn a multi-view semi-supervised graph with hierarchical attention for fraud detection. GDN~\cite{ding2021few} adopts a deviation loss to train GNN and uses a cross-network meta-learning algorithm for few-shot node anomaly detection. SemiADC~\cite{meng2021semi} simultaneously explores the time-series feature similarities and structure-based temporal correlations. TADDY \cite{liu2021taddy} constructs a node encoding to cover spatial and temporal knowledge and leverages a sole transformer model to capture the coupled spatial-temporal information. However, these methods are based on discrete-time snapshots of dynamical graphs that are not well suited to continuous time dynamical graphs, and the temporal information is simply added to the model using a linear network mapping as features, without taking into account temporal properties such as periodicity.

\section{PRELIMINARIES}
\subsection{Notations and Problem Formulation}
\textbf{Notations.} Continuous-time dynamic graphs are important for modeling relational data for many real-world complex applications. The dynamic graph can be represented as $\mathcal{G}=(\mathcal{V}, \mathcal{E})$, where $\mathcal{V} = {v_i}$ is the set of nodes involved in all temporal events, and $\mathcal{E}$ denotes a sequence of temporal events. Typically, let $\mathcal{E}=\{\delta(t_1), \delta(t_2),...,\delta(t_m)\}$ be an event stream that generates the temporal network and $m$ is the number of observed events, with an event $\delta(t) = (v_i, v_j, t, x_{ij})$ indicating an interaction or a link happens from source node $v_i$ to target node $v_j$ at time $t$, with associating edge feature $x_{ij}$. Note that there might be multiple links between the same pair of node identities at different timestamps.

\noindent\textbf{Problem formulation.}
In practical applications, it is very difficult to obtain large-scale labeled data due to the prohibitive cost of collecting such data. The goal of our work is to be able to improve the performance of anomaly detection based on very limited supervised knowledge. Taking the task of temporal event anomaly detection as an example, based on a dynamic graph $\mathcal{G}$, we assume that each event has a corresponding true label $y \in \mathcal{Y}^t$. However, due to the limitation of objective conditions, only a few samples of the label can be observed, denoted by $\mathcal{Y}^L$, and the rest are unlabeled data denoted by $\mathcal{Y}^U$. In our problem $|\mathcal{Y}^L| \ll |\mathcal{Y}^U|$, our task is to learn an anomaly score mapping function $\phi$ that separates abnormal and normal samples.

\begin{figure*}[t]
	\centering
	\includegraphics[width=\linewidth]{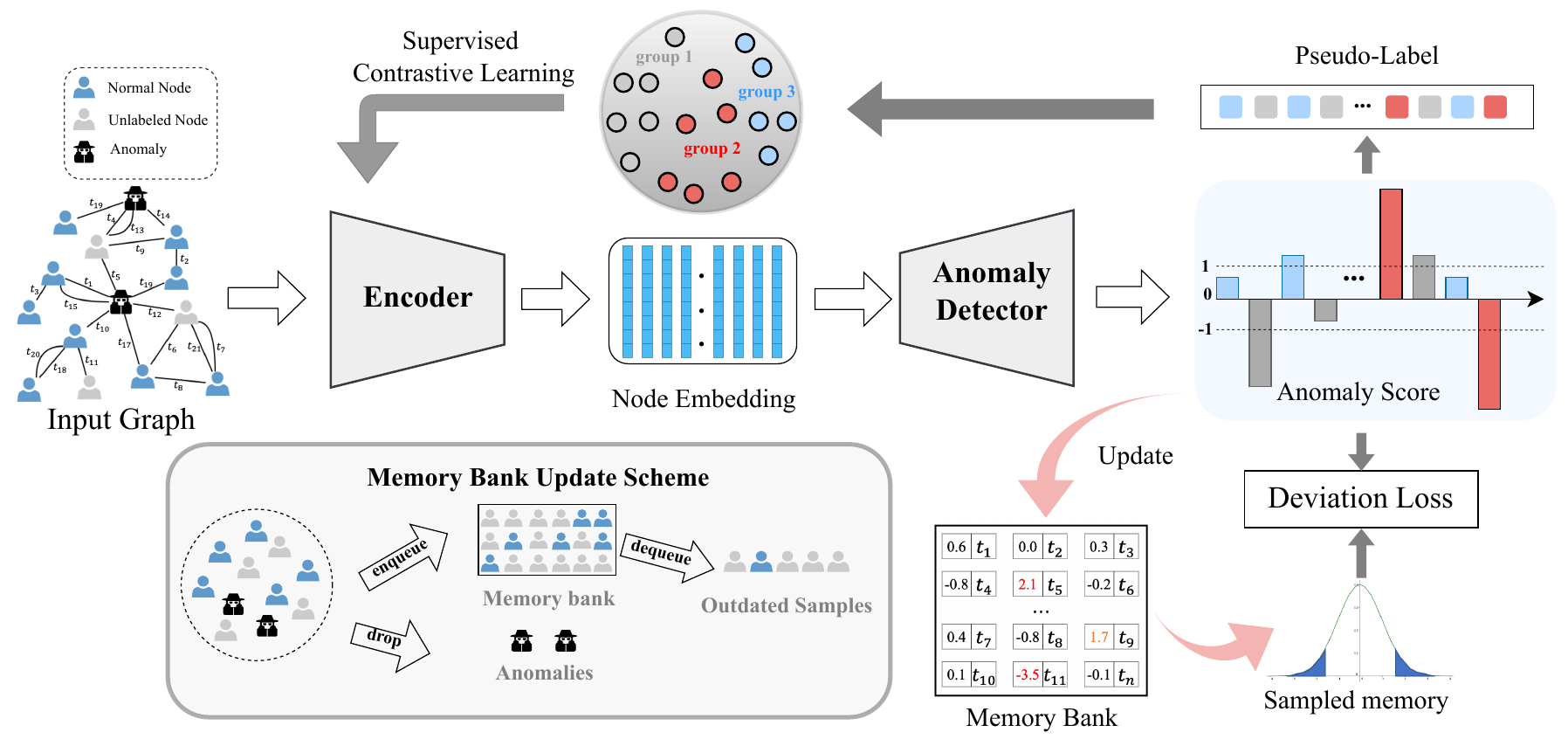}
 \vspace{-0.4cm}
	\caption{The proposed semi-supervised anomaly detection framework. SAD consists of four main components: the temporal graph encoder, the anomaly detector, the time-equipped memory bank, and the supervised contrastive learning module.}
	\label{fig:framework}
\vspace{-0.3cm}
\end{figure*}

\section{Method}
% 模型框图
In this section, we detail the components of the proposed framework SAD for anomaly detection on dynamic graphs. As shown in Figure~\ref{fig:framework}, the overall structure of SAD consists of the following main components: a temporal graph encoder that produces node embeddings (Section~\ref{sec:d_network}), the combination of an anomaly detector and a time-equipped memory bank to distinguish anomalous samples using prior knowledge of unlabeled data (Section~\ref{sec:d_network}), and a supervised contrastive learning module can yield pseudo-labeling to mine the potential of unlabeled data (Section~\ref{sec:cl}). We also detail several strategies for parameter learning under our framework (Section~\ref{sec:learning}).

\subsection{Deviation Networks with Memory Bank}
\label{sec:d_network}
To enable anomaly detection with few-shot labeled data, we construct a graph deviation network following~\cite{ding2021few}. However, for application to dynamic graphs, we first use a temporal graph encoder based on temporal encoding information to learn node representations and additionally construct a memory bank to dynamically record the overall statistical distribution of normal and unlabeled samples as a reference score to calculate the deviation loss that optimizes the model with few-shot labeled samples. The entire deviation network is semi-supervised, with abundant unlabeled samples used as statistical priors in the memory bank and few-shot labeled data used for calculating deviation losses.

\noindent\textbf{Temporal graph encoder.} Since dynamic graphs contain information about the time differences of edges, we first deploy a temporal graph encoder as a node embedding module to learn the topology of the graph by mapping nodes to a low-dimensional latent representation. The graph encoder is multiple GNN layers that take the graph $\mathcal{G}$ constructed at time $t$ as input and extract node representations $z_i(t)$ for all $v_i$ in $\mathcal{G}$ through the message-passing mechanism. Formally, the forward propagation at $k$-th layer is described as follows:
\begin{equation}
	\label{eq:encoder}
	\begin{aligned}
		 & \mathbf{h}_{N_i}^{(k)}(t)=\operatorname{AGG}^{(k)}\left(\left\{(\mathbf{h}_{j}^{(k-1)}(t), x_{ij}, \phi(\Delta t)): v_j \in \mathcal{N}(v_i,t)\right\}\right) \text {, } \\
		 & \mathbf{h}_{i}^{(k)}(t)=\operatorname{COMBINE}^{(k)}\left(\mathbf{h}_{i}^{(k-1)}(t), \mathbf{h}_{N_i}^{(k)}(t)\right),
	\end{aligned}
\end{equation}
where $\mathbf{h}_{i}^{(k)}(t)$ is the intermediate representations of node $v_i$ in the $k$-th layer at the timestamp $t$ and $v_j \in \mathcal{N}(v_i,t)$ denotes the set of first-order neighboring nodes of the node $v_i$ occurred prior to $t$. $x_{ij}$ is the associating edge feature between $v_i$ and $v_j$, $\Delta t = t - t_{ij}$ represents the relative timespan of two timestamps, and $\phi(\cdot)$ is a relative time encoder based on the cosine transform mentioned in~\cite{xu2020tgat}, which implements a functional time projection from the time domain to the continuous space and helps to capture the periodic patterns on the dynamic graph. $\operatorname{AGG}(\cdot)$ is an aggregation function for propagating and aggregating messages from neighborhoods, followed by a $\operatorname{COMBINE}(\cdot)$ function for combining representations from neighbors and its previous-layer representation. Note that the $\operatorname{AGG}(\cdot)$ and $\operatorname{COMBINE}(\cdot)$ functions in our framework can be flexibly designed for different applications without any constraints. Finally, we can obtain the node embedding $z_i(t) = \mathbf{h}_{i}^{(K)}(t) $ of node $v_i$ with the $K$-th aggregation layer. In practice, across all tasks, we adopt TGAT as the temporal graph encoder.

\noindent\textbf{Anomaly Detector.} 
To distinguish abnormal samples from normal samples, we introduce an anomaly detector to map the learned node embedding $z_i(t)$ to an anomaly score space, where normal samples are clustered and abnormal samples deviate from the set. Specifically, we use a simple feed-forward neural network $f_{\theta_a}(\cdot)$ as the anomaly detector:
\begin{equation}
	\label{eq:detector}
		 s_i(t) = f_{\theta_a}(z_i(t)) = \mathbf{W_2}\cdot \operatorname{ReLU}(\mathbf{W_1} \cdot z_i(t) + \mathbf{b_1}) + \mathbf{b_2},
\end{equation}
where $\mathbf{W_1}$, $\mathbf{b_1}$, $\mathbf{W_2}$ and $\mathbf{b_2}$ are learnable parameters of the anomaly detector. In our framework, $s_i(t)$ is a one-dimensional anomaly score with a value range from $-\infty$ to $+\infty$. Ideally, the anomaly scores of normal samples are clustered in a certain interval, and it is easy to find out the anomalies (outliers) based on the overall distribution of the anomaly scores.

\noindent\textbf{Memory bank.}
In essence, the task of SAD is to discriminate anomalous nodes from large-scale normal/unlabeled nodes in the dynamic graph utilizing a few labeled samples. The anomaly scores of normal nodes are clustered together in the anomaly score space after training with the optimization objective (which will be introduced later), while the anomaly scores of the anomaly samples will deviate from the overall distribution. Hence, we generate this distribution by using a memory bank $mem$ to record the historical anomaly scores of unlabeled and normal samples, based on the assumption that most of the unlabeled data are normal samples. The messages $m$ that need to be stored in the memory bank are described as the following:
\begin{equation}
   \begin{aligned}
	\label{eq:memory1}
		 m = s_i(t) ,\; \; \text{if $\;y_i(t) = 0$ or $-1$} 
         \end{aligned}
\end{equation}

\noindent where $y_i(t)$ is the label information of node $v_i$ at timestamp $t$ and its value $-1$ and $0$ represent that the node is an unlabeled or normal sample at the current timestamp, respectively.

Note that the attribute of the nodes in the dynamic graph evolves continuously, with consequent changes in the labels and anomaly scores. Perceptually, samples with longer time intervals should have less influence on the current statistical distribution, so we also record the corresponding time t for each anomaly score in the memory bank.
\begin{equation}
        \begin{aligned}
	\label{eq:memory2}
		 m = (s_i(t), t) ,\; \; \text{if $\;y_i(t) = 0$ or $-1$}
         \end{aligned}
\end{equation}

Meanwhile, the memory bank $mem$ is designed as a first-in-first-out (FIFO) queue of size $M$.
The queue size is controlled by discarding samples that are sufficiently old since outdated samples yield less gain due to changes in model parameters during the model training process.

\noindent\textbf{Deviation loss.} Inspired by the recent success of deviation networks~\cite{pang2019deep,ding2021few}, we adopt deviation loss as our main learning objective to enforce the model to separate out normal nodes and abnormal nodes whose characteristics significantly deviate from normal nodes in the anomaly score space. Different from their methods, we use a memory bank to produce a statistical distribution of normal samples instead of using a standard normal distribution as the reference score. This takes into account two main factors, one is that statistical distribution generated by real anomaly score can better show the properties of different datasets, and the other is that in dynamic graphs, the data distribution can fluctuate more significantly due to realistic situations such as important holidays or periodic events, so it is not suitable to use a simple standard normal distribution as the reference distribution of normal samples. 

To obtain the reference score in the memory bank, we assume that the distribution of the anomaly score satisfies the Gaussian distribution, which is a very robust choice to fit the anomaly scores of various datasets~\cite{Kriegel2011Interpreting}. Based on this assumption, we randomly sample a set of $M_s$ examples from the memory bank for calculating the reference score, which helps to improve the robustness of the model by introducing random perturbations. Meanwhile, we introduce the effect of time decay and use the relative timespan between the timestamp $t$ and the anomaly score storage time $t_i$ to calculate the weighting term for each statistical example $w_i(t) = \frac{1}{\ln (t-t_i + 1) + 1}$. The reference scores (mean $\mu_r(t)$ and standard deviation $\sigma_r(t)$) of the normal sample statistical distribution at the timestamp $t$ are calculated as follows:
\begin{equation}
        \begin{aligned}
	\label{eq:dev_ref}
		 \mu_r(t) &= \frac{1}{k}\sum_{i=1}^k w_i(t) \cdot r_i
        \\
            \sigma_r(t)  &= \sqrt{ \frac{\sum_{i=1}^k w_i(t)\cdot ( r_i - \mu_r(t))^2 }{k-1}}
         \end{aligned}
\end{equation}

With the reference scores at the timestamp $t$, we define the deviation $dev(v_i, t)$ between the anomaly score of node $v_i$ and the reference score in the following:
\begin{equation}
	\label{eq:dev_score}
		 dev(v_i, t) = \frac{s_i(t) - \mu_r(t)}{\sigma_r(t)}.
\end{equation}

The final deviation loss is then obtained by plugging deviation $dev(v_i, t)$ into the contrast loss~\cite{Hadsell2006Dimensionality} as follows:
\begin{equation}
	\label{eq:dev_loss}
		 \mathcal{L}^{dev} = (1-y_i(t))\cdot |dev(v_i, t)| + y_i(t) \cdot \max(0, m-|dev(v_i,t)|),
\end{equation}
where $y_i(t)$ is the label information of node $v_i$ at timestamp $t$ and $m$ is equivalent to a Z-Score confidence interval parameter. Note that in this semi-supervised task, only labeled samples are directly involved in computing the objective function. Optimizing this loss will enable the anomaly score of normal samples as close as possible to the overall distribution in the memory bank, while the anomaly scores of anomalies are far from the reference distribution.

\subsection{Contrastive Learning for Unlabeled Samples}
\label{sec:cl}
To fully exploit the potential of unlabeled samples, we generate a pseudo-label $\hat{y}_i(t)$ for each sample based on the existing deviation scores $dev(v_i, t)$ and involve them in the training of the graph encoder network. Here, we design a supervised contrastive learning task such that nodes with closer deviation scores also have more similar node representations, while nodes with larger deviation score differences have larger differences in their corresponding representations. The goal of this task is consistent with our deviation loss, which makes the difference between normal and anomaly samples deviate in both the representation space and anomaly score space. For a batch training sample with batch size $N$, we first obtain the deviation distance $\triangle d_{ij}= |dev(v_i, t_i) - dev(v_j, t_j)|$ between any two samples $v_i$ and $v_j$. On the standard normal distribution anomaly space, we group nodes with intervals less than $1 \sigma$ into the same class. Thus, for a single sample $v_i$, its supervised contrastive learning loss takes the following form:
\begin{equation}
    \begin{split}
	\label{eq:sup_loss}
 \mathcal{L}_i^{scl} & = \frac{-1}{N-1}\sum_{j,j\neq i}^N 1_{\triangle d_{ij}<1} \cdot \frac{1}{1+\triangle d_{ij}} \cdot \\ &   \ln  \frac{exp(z_i(t_i) \cdot z_j(t_j)/\tau)}{\sum_{k,i\neq k}^N exp(z_i(t_i) \cdot z_k(t_k)/\tau)} 
    \end{split}
\end{equation}
where $\tau$ is a scalar temperature parameter, and $\frac{1}{1+\triangle d_{ij}}$ denotes similarity weights between positive sample pairs. The final objective function is the sum of the losses of all nodes: $\mathcal{L}^{scl} = \sum^N \mathcal{L}_i^{scl}$.

\subsection{Learning procedure}
\label{sec:learning}
For anomaly detection tasks, the optimization of the whole network is based on deviation loss and contrastive learning loss. Noted that the contrastive learning loss is only used to train the parameters of the temporal graph encoder. Specifically, let $\theta_{enc}$ denote all the parameters of the temporal graph encoder, and $\theta_{ano}$ denote the learnable parameters in the anomaly detector. The optimization task of our model is as follows:
\begin{equation}
\label{eq:optimization_ano}
 \mathop{\arg\min}\limits_{\theta_{enc},\theta_{ano}}  \mathcal{L}^{dev}(\theta_{enc},\theta_{ano}) + \alpha\mathcal{L}^{scl}(\theta_{enc}),
\end{equation}
where $\alpha$ is a hyperparameter for balancing the contribution of contrastive learning loss.

In addition to anomaly detection tasks, we also present an end-to-end learning procedure to cope with different downstream tasks. Taking the downstream task of node classification as an example, in order to obtain the score results for node classification, we use a simple multi-layer perception (MLP) as the projection network to learn the classification results based on the node representations after the graph encoder. Let $\theta_{pro}$ denote the learnable parameters in the projection network. Assume that the downstream supervised task with loss function $\mathcal{L}^{sup}(\theta_{pro})$, the optimization task of our model is as follows:
\begin{equation}
\label{eq:optimization}
 \mathop{\arg\min}\limits_{\theta_{enc},\theta_{ano},\theta_{pro}} \mathcal{L}^{sup}(\theta_{pro}) + \alpha\mathcal{L}^{dev}(\theta_{enc},\theta_{ano}) + \beta\mathcal{L}^{scl}(\theta_{enc}),
\end{equation}
where $\alpha$ and $\beta$ are hyperparameters for balancing the contributions of two losses.

\begin{table}[t]
\resizebox{\linewidth}{!}{
	\begin{tabular}{lcccc}
		\toprule
		\textbf{ }     & \textbf{Wikipedia}   	& \textbf{Reddit}	 & \textbf{Mooc}	    & \textbf{Alipay}
		  \\
		\midrule
            \#Nodes                   & 9,227              & 10,984            & 7,074                 & 3,575,301                \\
		  \#Edges                   & 157,474            & 672,447           & 411,749               & 53,789,768                \\
		  \#Edge features           & 172               & 172              & 4                   & 100                \\
            \#Anomalies            & 217                & 366               & 4,066                   & 24,979                \\
		  Timespan                  & 30 days           & 30 days          & 30 days                   & 90 days                \\
            Pos. label meaning    &posting banned     &editing banned    &dropping out          &fraudster          \\
            Chronological Split       &70\%-15\%-15\%     &70\%-15\%-15\%    &70\%-15\%-15\%        &70\%-15\%-15\%     \\
		\bottomrule
	\end{tabular}}
        \vspace{-0.2cm}
        \caption{Statistics of datasets.}
	\label{tab:datasets}
        \vspace{-0.3cm}
\end{table}

\section{Experiments}
\subsection{Experimental Setup}
\textbf{Datasets.} In this paper, we use four real-world datasets, including three public bipartite interaction dynamic graphs and an industrial dataset. Wikipedia~\cite{kumar2019predicting} is a dynamic network tracking user edits on wiki pages, where an interaction event represents the page edited by the user. Dynamic labels indicate whether a user is banned from posting. Reddit~\cite{kumar2019predicting} is a dynamic network tracking active users posting in subreddits, where an interaction event represents a user posting on a subreddit. The dynamic binary labels indicate if a user is banned from posting under a subreddit. MOOC~\cite{kumar2019predicting} is a dynamic network tracking students' actions on MOOC online course platforms, where an interaction event represents user actions on the course activity. The Alipay dataset is a dynamic financial transaction network collected from the Alipay platform, and dynamic labels indicate whether a user is a fraudster. Note that the Alipay dataset
has undergone a series of data-possessing operations, so it cannot represent real business information. For all tasks and datasets, we adopt the same chronological split with 70\% for training and 15\% for validation and testing according to node interaction timestamps. The statistics are summarized in Table \ref{tab:datasets}.

\noindent\textbf{Implementation details.} Regarding the proposed SAD model, we use a two-layer, two-head TGAT~\cite{xu2020tgat} as a graph network encoder to produce 128-dimensional node representations. For model hyperparameters, we fix the following configuration across all experiments without further tuning: we adopt Adam as the optimizer with an initial learning rate of 0.0005 and a batch size of 256 for both training. We adopt mini-batch training for SAD and sample two-hop subgraphs with 20 nodes per hop. For the memory bank, we set the memory size $M$ as 4000, and the sampled size $M_s$ as 1000. To better measure model performance in terms of AUC metrics, we choose the node classification task (similar to anomaly node detection) as the downstream task, so we adopt Eq.\eqref{eq:optimization} as the model optimization objective and find that pick $\alpha=0.1$ and $\beta=0.01$ performs well across all datasets. The proposed method is implemented using PyTorch~\cite{pytorch} 1.10.1 and trained on a cloud server with NVIDIA Tesla V100 GPU.
Code is made publicly available at \url{https://github.com/D10Andy/SAD} for reproducibility.

\begin{table}[t]
        \centering
\resizebox{\linewidth}{!}{
	\begin{tabular}{l|cccc}
		\toprule
		 {\textbf{Methods}} & \textbf{Wikipedia}               & \textbf{Reddit}                   & \textbf{Mooc}  & \textbf{Alipay}
		  \\
		 \midrule
		 TGAT                         & 83.23 {$\pm$ 0.84}               & 67.06 {$\pm$ 0.69}                & 66.88 {$\pm$ 0.68}           & 92.53 {$\pm$ 0.93}                
   \\
   TGN                         & 84.67 {$\pm$ 0.36}               &62.66 {$\pm$ 0.85}                & 67.07 {$\pm$ 0.73}           & 92.84 {$\pm$ 0.81}                
   \\
		                                             Radar                         & 82.91 {$\pm$ 0.97}               & 61.46 {$\pm$ 1.27}                & 62.14 {$\pm$0.89}           & 88.18 {$\pm$ 1.05} 
    \\
		                                             DOMINANT                         & 85.84 {$\pm$ 0.63}               & 64.66 {$\pm$ 1.29}                &  65.41{$\pm$ 0.72}           & 91.57 {$\pm$ 0.93}    
                                               \\
		                                             SemiGNN                         & 84.65 {$\pm$ 0.82}               & 64.18 {$\pm$ 0.78}                & 64.98 {$\pm$0.63}           & 92.29 {$\pm$ 0.85}

  \\
		                                             GDN                         & 85.12 {$\pm$ 0.69}               & 67.02 {$\pm$ 0.51}                & 66.21 {$\pm$ 0.74}           & 93.64 {$\pm$ 0.79}                  \\
		                                             TADDY                     & 84.72 {$\pm$ 1.01}               & 67.95 {$\pm$ 0.94}                & 68.47 {$\pm$ 0.76}           & 93.15 {$\pm$ 0.88}          \\
		                                             SAD                             & \textbf{86.77 {$\pm$ 0.24} }     & \textbf{68.77 {$\pm$ 0.75} }      & \textbf{69.44 {$\pm$ 0.87} } & \textbf{94.48 {$\pm$ 0.65} } \\
		\bottomrule
	\end{tabular}}
    \vspace{-0.1cm}
    \caption{Overall performance of all methods in terms of AUC on dynamic node classification tasks. Means and standard deviations were computed over 10 runs.}
    \vspace{-0.3cm}
    \label{tab:overallperformance}
\end{table}

%the baseline consists of our backbone model TGAT (in a supervised fashion) and the state-of-the-art supervised learning method TGN, as well as five graph anomaly detection methods.

\subsection{Overall Performance}
We compare our SAD against several strong baselines of graph anomaly detection methods on node classification tasks, the baselines in our experiments include our backbone model TGAT and the state-of-the-art supervised learning method TGN-no-mem ~\cite{tgn2020}, as well as five graph anomaly detection methods, which are specifically divided into unsupervised anomaly detection methods (Radar and DOMINANT) and semi-supervised anomaly detection methods (SemiGNN, GDN, and TADDY). In order to be able to fairly compare the performance of the different methods, we use the same structure of the projection network to learn the classification results based on the node embedding outputted by their anomaly detection tasks. Table \ref{tab:overallperformance} summarizes the dynamic node classification results on different datasets. For all experiments, we report the average results with standard deviations of 10 different runs.

As shown in Table \ref{tab:overallperformance}, our approach outperforms all baselines on all datasets. Due to the lack of dynamic information, these static graph anomaly detection methods (Radar, SemiGNN, DOMINANT, and GDN) perform worse than the dynamic graph supervision model TGAT on Reddit and Mooc, which are more sensitive to temporal information. On these datasets with large differences in the proportion of positive and negative samples, our method shows a significant improvement compared to TGAT. Specifically, the relative average AUC value improvements on Wikipedia, Reddit, Mooc, and Alipay are $3.54\%$, $1.70\%$, $2.56\%$, and $1.95\%$, respectively. Compared with the recent dynamic graph anomaly detection method TADDY, our method improves $2.05\%$, $0.82\%$, $0.97\%$, and $1.33\%$ on four datasets, respectively. We believe the significant improvements are that SAD captures more information on time-decaying state changes and that the memory bank mechanism can effectively obtain the distribution of normal data, which helps to detect anomaly samples.

\begin{figure}[t]
	\centering
	\subfigure[Wikipedia]{\includegraphics[width=0.44\hsize, height=0.4\hsize]{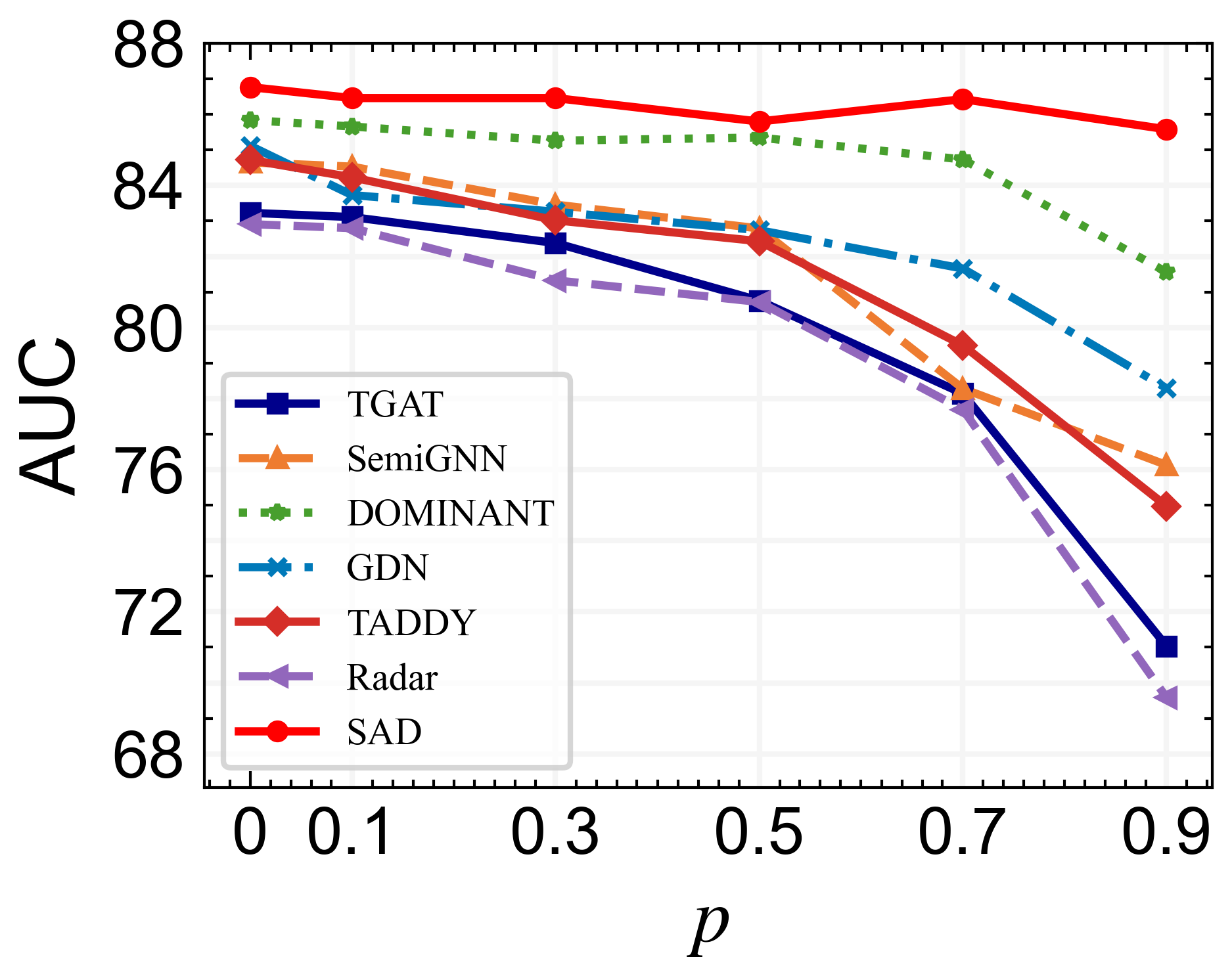}\label{over_a}}\hspace{0.2cm}
	\subfigure[Reddit]{\includegraphics[width=0.44\hsize, height=0.4\hsize]{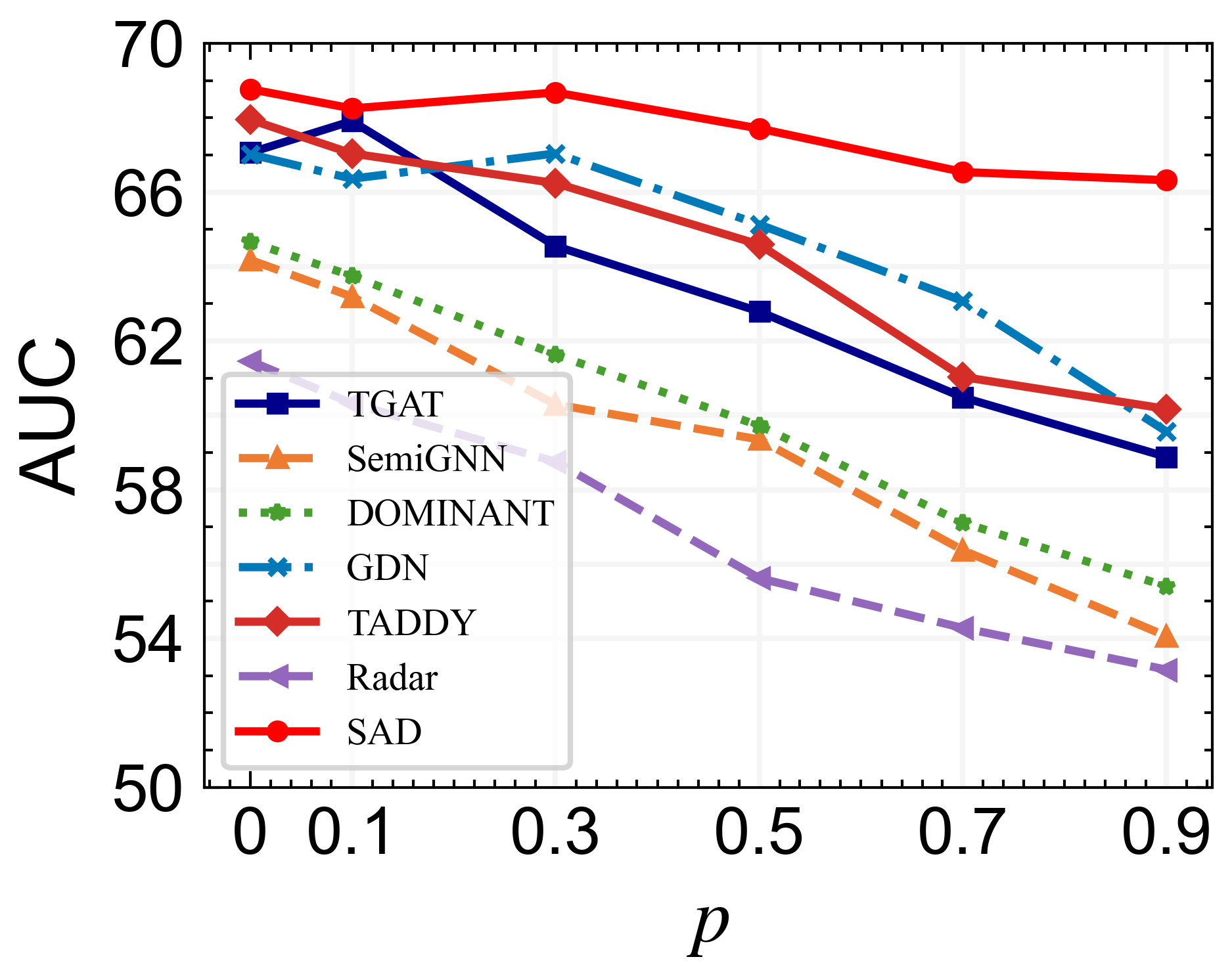}\label{over_b}}\hspace{0.2cm}
        \vspace{-0.1cm}
	\caption{Dynamic node classification task results under different drop ratios $p$ on Wikipedia and Reddit, respectively.}
	\label{fig:few-shot evaluation}
 \vspace{-0.1cm}
\end{figure}

\begin{figure*}[!ht]
	\centering
	\subfigure[TGAT]{\includegraphics[width=0.22\hsize, height=0.22\hsize]{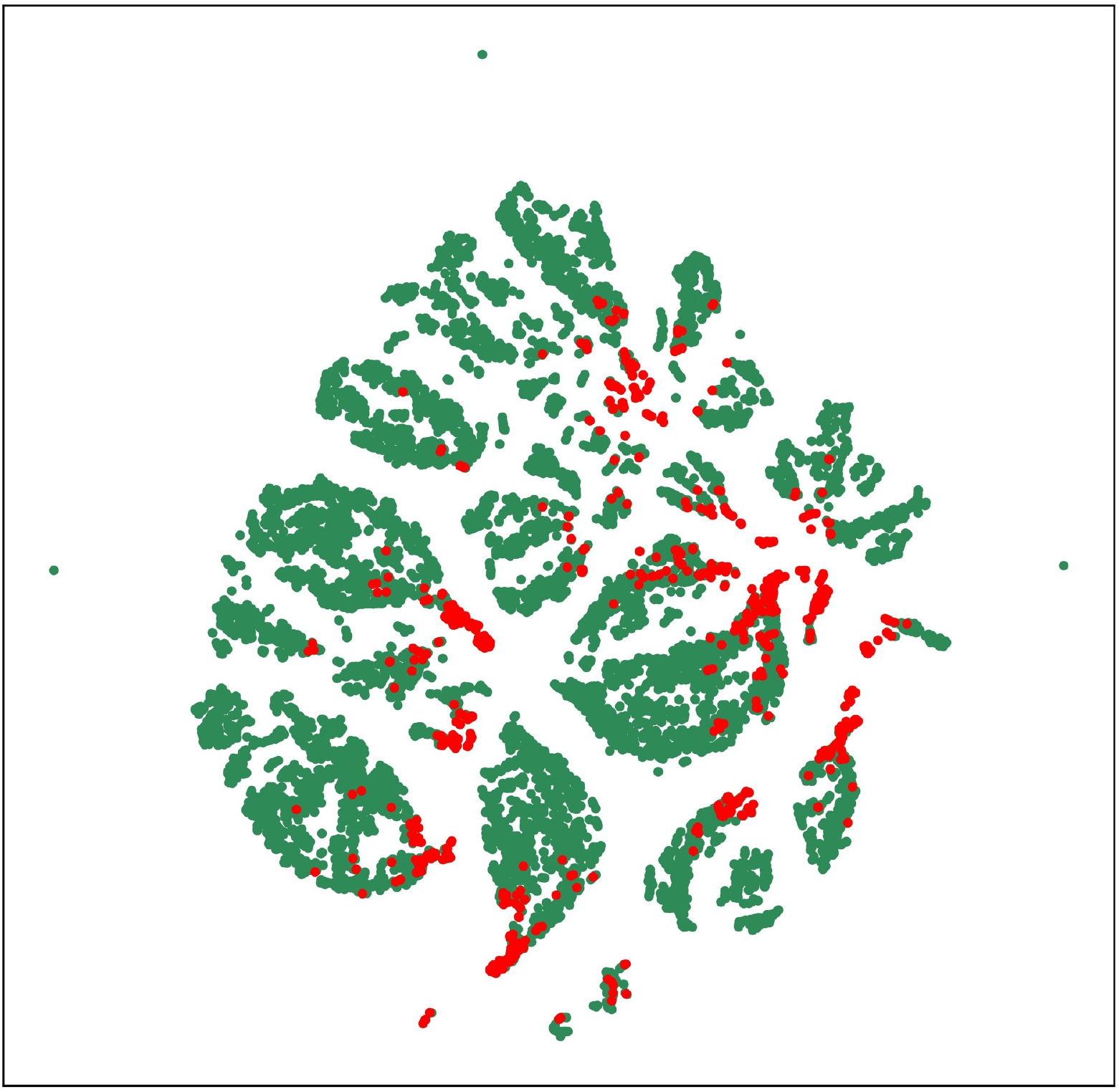}\label{b}}\hspace{0.2cm}
	\subfigure[GDN]{\includegraphics[width=0.22\hsize, height=0.22\hsize]{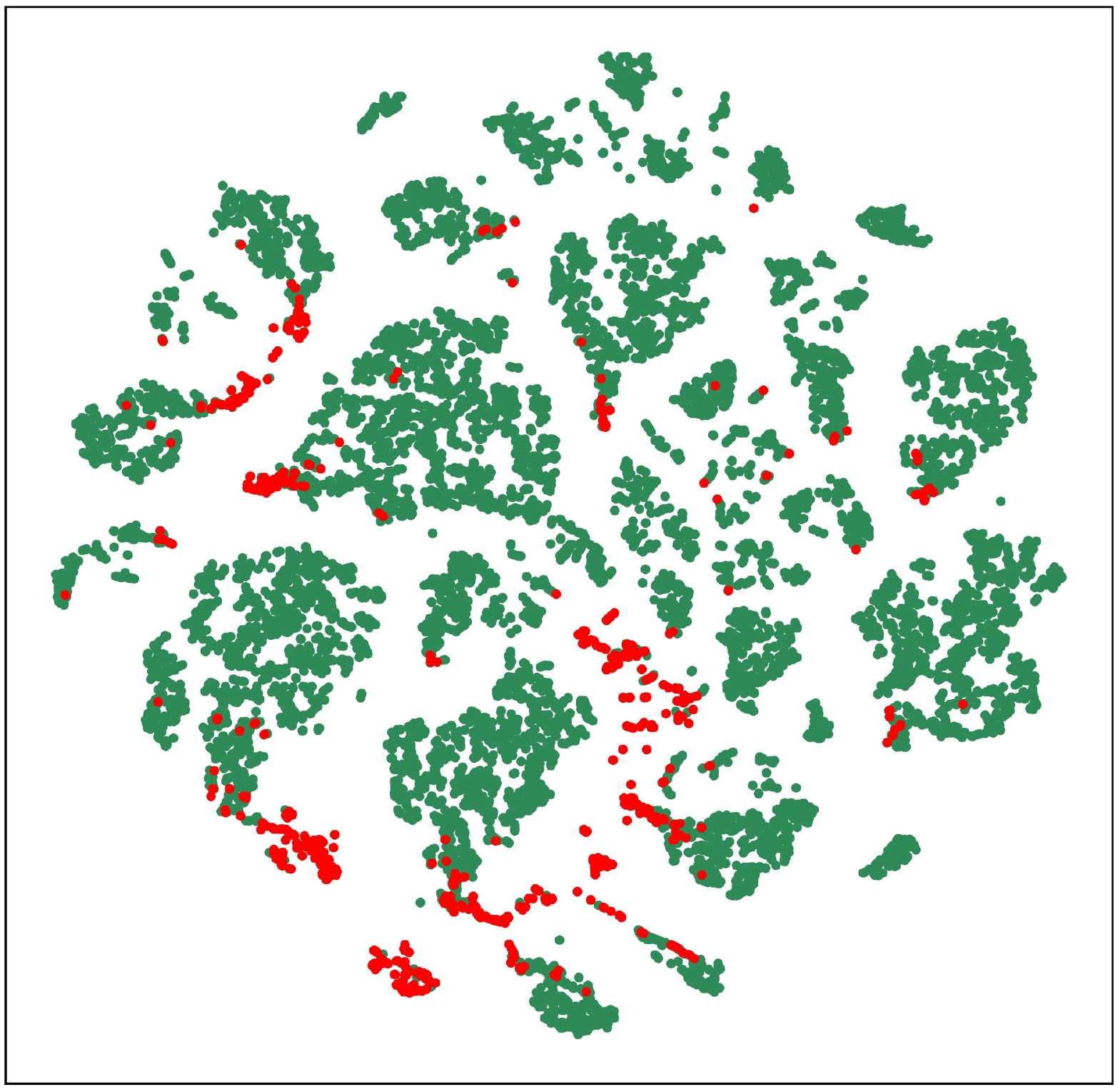}\label{d}}\hspace{0.2cm}
        \subfigure[TADDY]{\includegraphics[width=0.22\hsize, height=0.22\hsize]{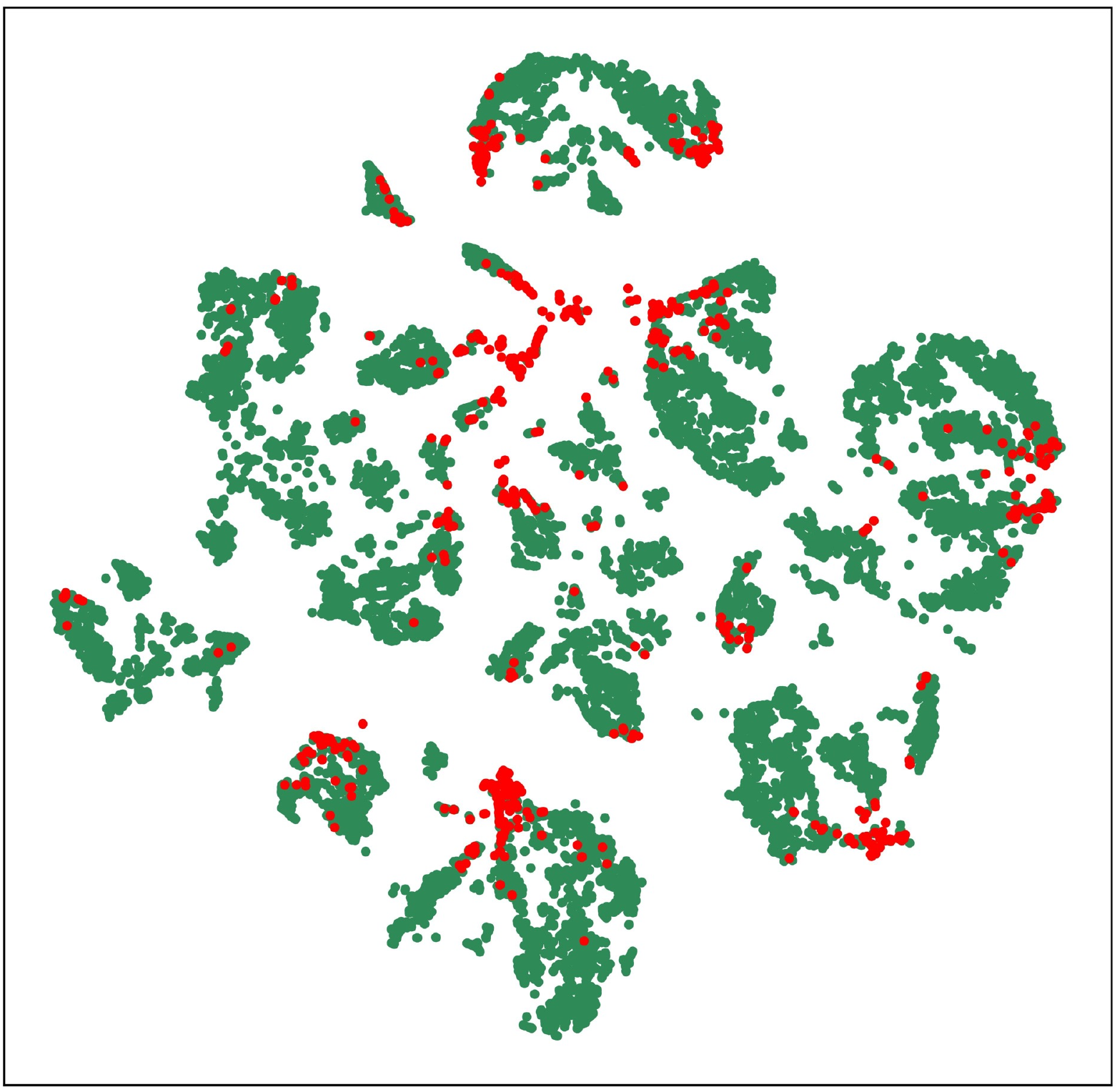}\label{c}}\hspace{0.2cm}
        \subfigure[SAD]{\includegraphics[width=0.22\hsize, height=0.22\hsize]{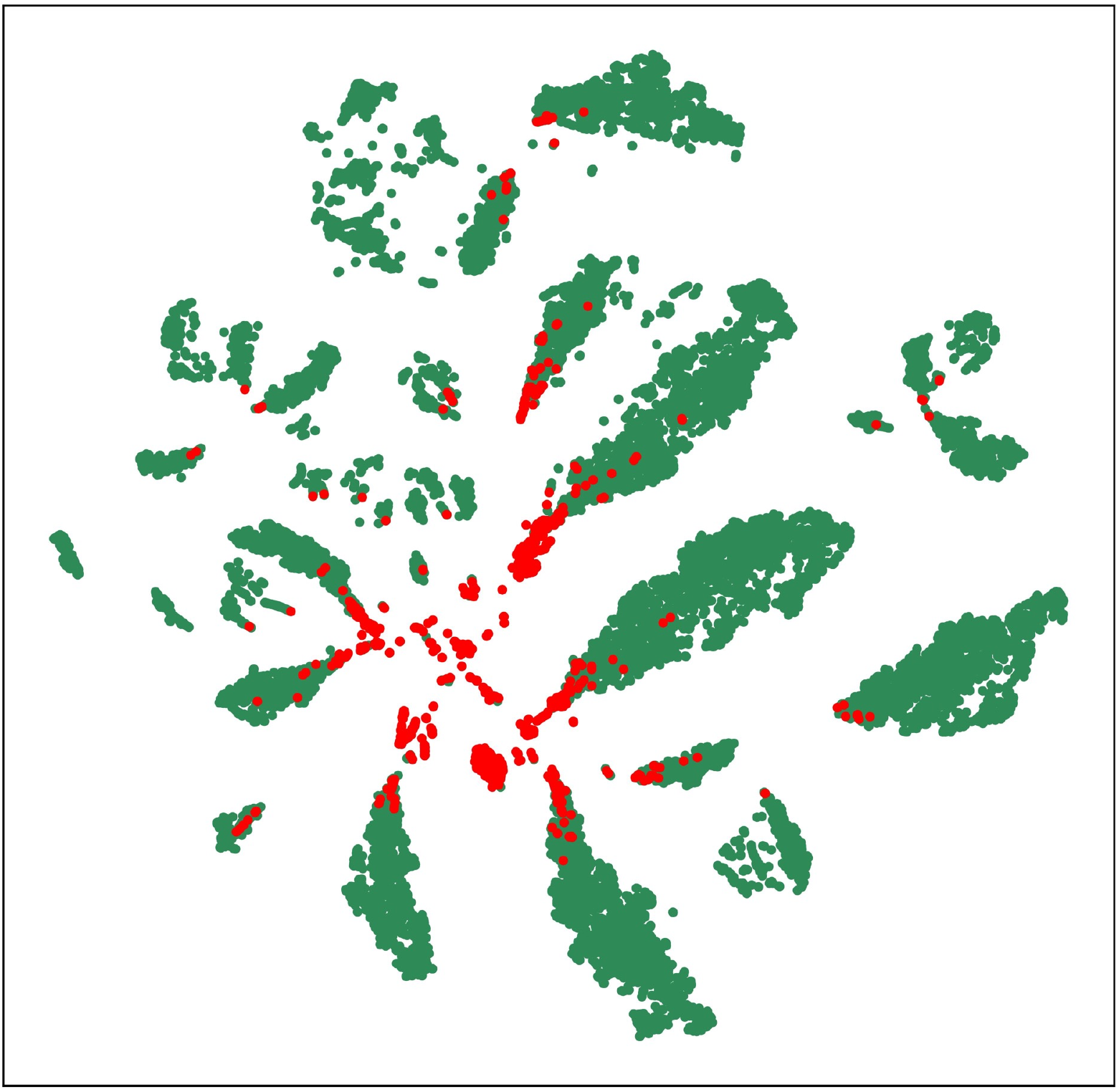}\label{a}}\hspace{0.2cm}
        \vspace{-0.1cm}
	\caption{Visualization of the learned node embeddings w.r.t. different methods on Alipay. The red and green points represent the abnormal and normal samples, respectively.}
	\label{fig:t-SNE visualization}
 \vspace{-0.1cm}
\end{figure*}

\subsection{Few-shot Evaluation}
The training samples used in the overall performance are annotated. However, in real application scenarios, annotated samples are always difficult to obtain, and anomaly detection algorithms need to be applied to distinguish anomalous nodes in large-scale unlabeled samples. To thoroughly evaluate the benefits of different graph anomaly detection approaches in few-shot scenarios, we varied the drop rate $p$, i.e., the ratio of randomly removed node labels in the training set, from $0.1$ to $0.9$ with a step size of $0.2$. 

The curve of AUC values of different models and datasets along with the label drop ratio is drawn in Figure \ref{fig:few-shot evaluation}. SAD achieves state-of-the-art performance in all cases across two dynamic datasets. Noticeably, we observe that SAD achieves optimal performance on both datasets when the drop rate is at $0.3$. The result indicates that there is a lot of redundant or noisy information in the current graph datasets, which easily leads to overfitting of the model training. By dropping this part of label information and using pseudo-labeled data to participate in the training, the performance of the model is improved instead. Other methods have a significant degradation in performance at a drop ratio of $0.5$. Nevertheless, SAD still outperforms all the baselines by a large margin on two datasets, and the downstream performance is not significantly sacrificed even for a particular large dropping ratio (e.g., 0.7 and 0.9). This indicates that even for few-shot scenarios, SAD could still mine useful statistical information for downstream tasks and distinguish anomalous nodes with information from abundant unlabeled samples. Overall, we believe that the performance improvement in our model comes from two reasons: one is that the memory bank-dominated anomaly detector allows the model to learn a discriminative representation between normal and anomalous samples, and the other is that contrastive learning based on pseudo labels helps to learn generic representations of nodes even in the presence of large amounts of unlabeled data.

\subsection{Visualization Analysis}
In this part, we visualize the learned node embeddings to evaluate the proposed method. And the 2D t-SNE~\cite{maaten2008visualizing} is used to reduce the dimensions of the hidden layer output node embeddings from 128 to 2. Figure \ref{fig:t-SNE visualization} shows the visualization of node embeddings learned by TGAT, GDN, TADDY, and SAD on Alipay. As we can see, the node embeddings of SAD are able to separate the abnormal and normal samples well. The supervised learning method TGAT is prone to overfitting during training on datasets with class imbalance, resulting in poor discrimination of node representation learning. GDN is not effective in learning node embedding because it does not take into account dynamic information and historical data statistical distributions. TADDY is based on modeling discrete-time snapshots of dynamical graphs that are not well suited to continuous-time dynamical graphs. Overall, these results show that the proposed method can learn useful node embeddings for anomaly detection tasks on dynamic graphs.

\subsection{Ablation Study}
Finally, we conduct an ablation study to better examine the contribution of different components in the proposed framework, detailed as follows:
%namely the deviation network, the memory bank mechanism, the time decay technique in the memory bank, and the supervised contrastive learning module, detailed as follows:
%\vspace{-0.1cm}
\begin{itemize}
    \item \textbf {backbone.}  This entity indicates that only the graph coding network is used for dynamic node classification tasks in a supervised learning manner.
    %\vspace{-0.1cm}
    \item \textbf {w/dev.}  This variant indicates that the deviation network based on standard normal distribution is used on the basis of the backbone model, without the memory bank mechanism.
    %\vspace{-0.1cm}
    \item \textbf {w/mem.} This variant indicates that the memory bank mechanism without the time decay technique is used on the basis of the w/dev model.
    %\vspace{-0.1cm}
    \item \textbf {w/time.} This variant indicates that the time decay technique in the memory bank is used on the basis of the w/mem model.
    %\vspace{-0.1cm}
    \item \textbf {w/scl.} This variant indicates that the supervised contrastive learning module is used on the basis of the w/time model, which corresponds to our full model.
\end{itemize}

\begin{table}[t]
	\begin{tabular}{lccc}
		\toprule
		\textbf{ } & \textbf{Wikipedia}               & \textbf{Reddit}                   & \textbf{Mooc}  	\\
		\midrule
		TGAT                         &80.76 {$\pm$ 2.30}               & 62.79 {$\pm$ 3.42}                & 64.04 {$\pm$ 1.02}                        \\
		\midrule
		w/dev                        &82.45 {$\pm$ 0.64}               & 64.15 {$\pm$ 2.93}                & 65.33 {$\pm$ 1.67}                           \\
		w/mem                        &85.20 {$\pm$ 1.30}               & 66.96 {$\pm$ 1.51}                & 67.25 {$\pm$ 0.75}                            \\
		  w/time                     &85.44 {$\pm$ 0.75}               & 66.78 {$\pm$ 1.98}                & 67.53 {$\pm$ 0.93}                   \\
           % \midrule
		w/scl                             &\textbf{85.80 {$\pm$ 1.32} }     &\textbf{67.71 {$\pm$ 0.75} }      &\textbf{67.57 {$\pm$ 0.54} } \\
		\bottomrule
	\end{tabular}
 \vspace{-0.1cm}
    \caption{Results of the ablation study on the dynamic node classification task under the label-dropping ratio of $0.5$. 
    % Mean and standard deviations are computed over 10 runs.
    }
        \vspace{-0.1cm}
	\label{tab:ablation study}
\end{table}
\vspace{-0.1cm}

To verify the performance of the model in a large number of unlabeled sample scenarios, we manually drop $0.5$ of the labels of the training set at random. Table \ref{tab:ablation study} present the results given by our several sub-modules. We summarize our observations as follows: Firstly, each of the incremental submodules proposed in our paper helps to improve the performance of the model. Among them, the memory bank mechanism yields the largest performance improvements, e.g., $2.75\%$, $2.81\%$, and $1.92\%$ performance improvement on Wikipedia, Reddit, and Mooc, respectively. Secondly, the complete model outperforms the base model on the three dynamic datasets by $5.04\%$, $4.92\%$, and $3.17\%$, respectively, which verifies the effectiveness of the proposed submodule by introducing the deviation loss and contrastive learning based on pseudo labels.

\section{Conclusion}
We present a semi-supervised learning framework SAD for detecting anomalies over dynamic graphs. The proposed framework utilizes a temporal graph network and an anomaly detector to learn the anomaly score and uses a time-equipped memory bank to record the overall statistical distribution of normal and unlabeled samples as prior knowledge to guide the subsequent learning of the model in a semi-supervised manner. To further explore the potential of unlabeled samples, we produce pseudo-labels based on the difference between the anomaly scores and use these labels directly to train the backbone network in a supervised contrastive learning manner. Our results from extensive experiments demonstrate that the proposed SAD can achieves competitive performance even with a small amount of labeled data.

\bibliographystyle{named}
\bibliography{ijcai22}

\end{document}